\documentclass{article}

\usepackage{amssymb,latexsym}

\usepackage[pdftex]{graphicx}

\usepackage{hyperref}
\usepackage{xcolor}
\hypersetup{
    colorlinks,
    linkcolor={red!50!black},
    citecolor={blue!50!black},
    urlcolor={blue!80!black}
}

\begin{document}

\pagestyle{plain}

\newtheorem{theorem}{Theorem}[section]
\newtheorem{proposition}[theorem]{Proposition}
\newtheorem{lemma}[theorem]{Lemma}
\newtheorem{corollary}[theorem]{Corollary}
\newtheorem{definition}[theorem]{Definition}
\newtheorem{remark}[theorem]{Remark}
\newtheorem{exempl}{Example}[section]

\newenvironment{example}{\begin{exempl}  \em}{\hfill $\square$

\end{exempl}}  \vspace{.5cm}

\renewcommand{\contentsname}{ }

\title{chemSKI with tokens: world building and economy in the SKI universe}
\author{Marius Buliga \\ 
\\
Institute of Mathematics, Romanian Academy \\
P.O. BOX 1-764, RO 014700\\
Bucure\c sti, Romania\\ 
{\footnotesize Marius.Buliga@imar.ro , mbuliga@protonmail.ch}}

\date{01.06.2023}

\maketitle

\begin{abstract}
chemSKI with tokens is a confluent graph rewrite system where all rewrites are local, which moreover can be used to do SKI calculus reductions. The graph rewrites of chemSKI are made conservative by the use of tokens. We thus achieve several goals: conservative rewrites in a chemical style, a solution to the problem of new edge names in a distributed, decentralized graphical reduction and a new estimation of the cost of a combinatory calculus computation.

This formalism can be used either as an artificial chemistry or as a model of a virtual decentralized machine which performs only local reductions.
\end{abstract}

\section{Introduction}
\label{Introduction}

In the article \cite{chemski1} \href{https://mbuliga.github.io/chemski/chemski.html}{chemSKI and chemlambda}  I introduced the  purely local graph rewrite system chemSKI for the SKI combinators calculus. I was motivated to build chemSKI because in the article \cite{krushewski}  Combinatory Chemistry: Towards a Simple Model of Emergent Evolution \href{https://arxiv.org/abs/2003.07916}{arXiv:2003.07916}  Kruszewski and Mikolov asked for a graph-rewriting version of their  term rewrite system in chemlambda style. 

chemSKI is not a graph rewrite version of Combinatory Chemistry, though. Instead is a confluent graph rewrite system where all rewrites are local, which moreover can be used to do SKI calculus reductions. This is true in the same sense that Lafont Interaction Combinators \cite{lafont-comb} can be used to do lambda calculus reductions. That is because there is a parser from a SKI term to a chemSKI graph and a partial inverse function which can associate a SKI term to a free edge of a chemSKI graph, such that term reductions in SKI calculus correspond to sequences of graph rewrites in chemSKI.

Here I explain chemSKI \cite{chemskirepo}  \href{https://github.com/mbuliga/chemski}{[chemSKI repository]} with tokens, where the graph rewrites of chemSKI are made conservative by the use of tokens. We thus achieve two goals: conservative rewrites in a chemical style, and a new estimation of the cost of a computation. 

\tableofcontents

\section{Graphs and nodes of chemSKI}
\label{GraphsAndNodes}

The graphs are called "molecules" and they are formed by nodes connected by edges. Each node has a type and 1, 2 or 3 numbered node ports. The types of nodes of chemSKI are S, K, I, A, Arrow, FRIN, FROUT, see \cite{chemskirepo} \href{https://github.com/mbuliga/chemski/blob/master/js/nodes.js#L97}{[nodes.js line 21]} (contains the all types of nodes used in the chemlambda project).

\vspace{.5cm}
 
\centerline{\includegraphics[width=0.9\textwidth]{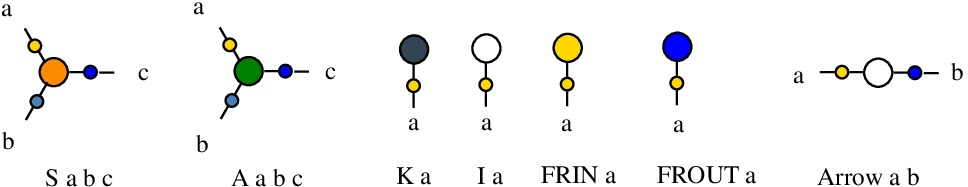}}
\vspace{.5cm}

There are two 1-valent nodes, for the combinators I and K, and two 3-valent nodes, one for the combinator S and simultaneously for fanout, the other for the application (A like in \cite{chemlambdahistory} \href{https://mbuliga.github.io/quinegraphs/history-of-chemlambda.html#ChemlambdaV2}{chemlambda v2}). There are also two 1-valent nodes, FRIN (free in) and FROUT (free out) and a 2-valent node Arrow which serve the same roles as in chemlambda.

As explained in \cite{buligachemlambda} \href{https://arxiv.org/abs/2003.14332}{arXiv.2003.14332}{arXiv.2003.14332}, we use the mol notation, for example a 3-valent node S is described by the string "S a b c" which means that the node type is "S", the port 1 of the node is linked to the edge named "a", the port 2 of the node is linked to the edge named "b" and finally the port 3 of the node is linked to the edge named "c".

\vspace{.5cm}
 
\centerline{\includegraphics[width=0.9\textwidth]{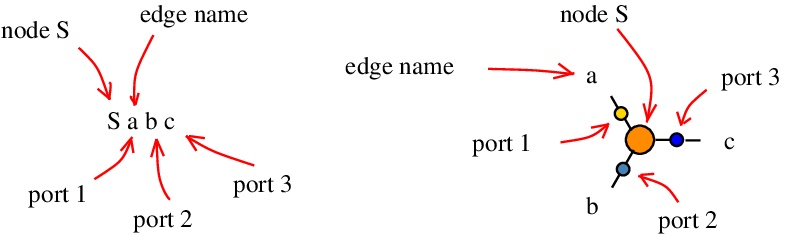}}
\vspace{.5cm}

 The edge names are arbitrary strings (with letters a-z, A-Z, 0-9, without space character) and different edges of the graph have different names. To each node port there is only one edge connected. They follow the same conventions as \cite{chemlambdahistory} \href{https://mbuliga.github.io/quinegraphs/history-of-chemlambda.html#ChemlambdaV2}{chemlambda v2}.

The color codes for node types and ports \cite{chemskirepo} \href{https://github.com/mbuliga/chemski/blob/master/js/nodes.js#L153}{[nodes.js line 153]} are described in the next figure.

\vspace{.5cm}
 
\centerline{\includegraphics[width=0.9\textwidth]{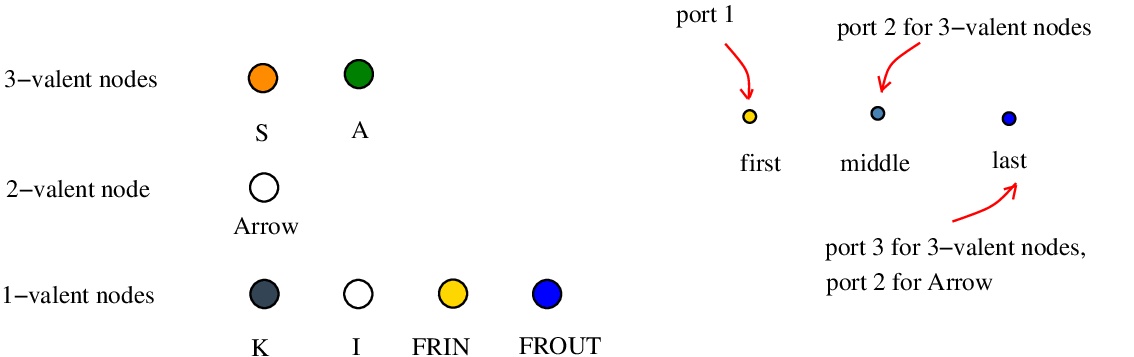}}
\vspace{.5cm}

FRIN and FROUT are used to cap the free half-edges of the graph, if any.  This is done according to the type of the ports of nodes: the first port is always "in", the last port is always "out" and, for the middle ports of the two trivalent nodes the rules are: for the node S the middle port is of type "out", for the node "A" the middle port is of type "in". in the following figure we see how or and

\vspace{.5cm}
 
\centerline{\includegraphics[width=0.9\textwidth]{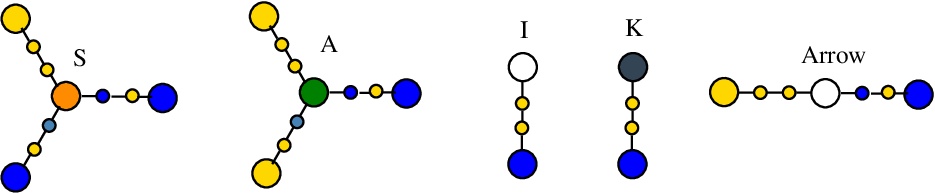}}
\vspace{.5cm}

The Arrow node was introduced to allow simultaneous application of rewrites, followed by the deletion of the introduced Arrow nodes by the rewrite COMB (which "combs" the graph). Together with FRIN and FROUT, they are used in all chemlambda rewrite systems.

It is possible to have edges connected to the same node, but at different ports. For example "S b b a" describes a node of type "S", with an edge named "b" connecting port 1 and port 2 of that node.

\vspace{.5cm}
 
\centerline{\includegraphics[width=0.9\textwidth]{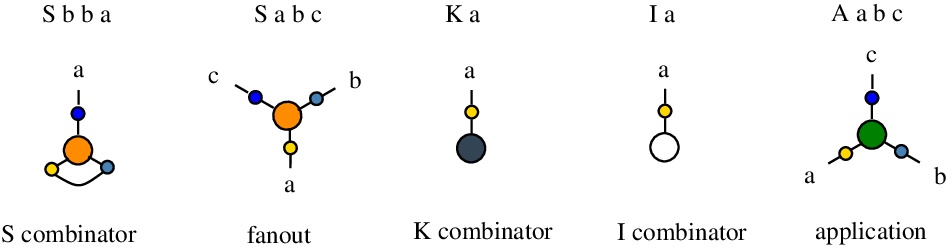}}
\vspace{.5cm}

  The rewrites concerning the node S are different, according to the existence of an edge between S node ports 1 and 2 (in this case S represents the combinator S) or not (in this case S is a fanout node).

Any molecule (ie chemSKI graph) is described by a mol file. A mol file is a list of nodes, where each node is a list which starts with the node type and continues with the edge names connected to ports, in the order of the ports. Therefore, as a string, a mol file needs a line separator (between node lines) and a field separator (within each line). As an example, the molecule which is the translation of the combinator ((S K) K) I is described by the mol file

$\left.\right.$ \newline $\left.\right.$ \newline
FROUT 1\newline
S 2 2 3\newline
A 3 4 5\newline
K 4\newline
A 5 6 7\newline
K 6\newline
A 7 8 1\newline
I 8

$\left.\right.$ \newline
with space "  " as field separator and new line as a line separator. In the simulation section  \href{https://mbuliga.github.io/chemski/chemski-with-tokens.html#PlayPlace}{[Play with chemSKI section]} we use " \^{ } " as the line separator, for example the same molecule appears as the string

$\left.\right.$ \newline $\left.\right.$ \newline
FROUT 6 \^{ } S 10 10 1 \^{ } A 1 3 2 \^{ } K 3 \^{ } A 2 5 4 \^{ } K 5 \^{ } A 4 7 6 \^{ } I 7 \newline

$\left.\right.$ \newline
Mind that the edge names are different, which is not important, because they are arbitrary.

The line separator is not important, because we can deduce the number of ports (ie the number of edge names on a node line) from the node type, therefore we shall also use a notation like

$\left.\right.$ \newline $\left.\right.$ \newline
FROUT 6 S 10 10 1 A 1 3 2 K 3 A 2 5 4 K 5 A 4 7 6 I 7$\left.\right.$ \newline

$\left.\right.$ \newline
for the same molecule, where the line and field separator are the same. Thus a mol file appears as a string of node types and edge names. This convention is useful in the section about rewrites.

\section{Conservative rewrites with tokens}
\label{ConservativeTokens}

Each chemSKI rewrite, section \ref{ChemskiRewrites},  is described by a left hand side (LHS) and a right hand side (RHS) pattern, which are mol files. There is also a chemical interpretation, where we see each rewrite as a chemical reaction, which transforms the LHS into RHS. In order to be conservative in nodes and edges, we need to add tokens, ie a finite number (up to isomorphism) of small graphs.

We shall use the following tokens  \cite{chemskirepo} \href{https://github.com/mbuliga/chemski/blob/master/js/nodes.js#L22}{[nodes.js line 22]}:

\vspace{.5cm}
 
\centerline{\includegraphics[width=0.9\textwidth]{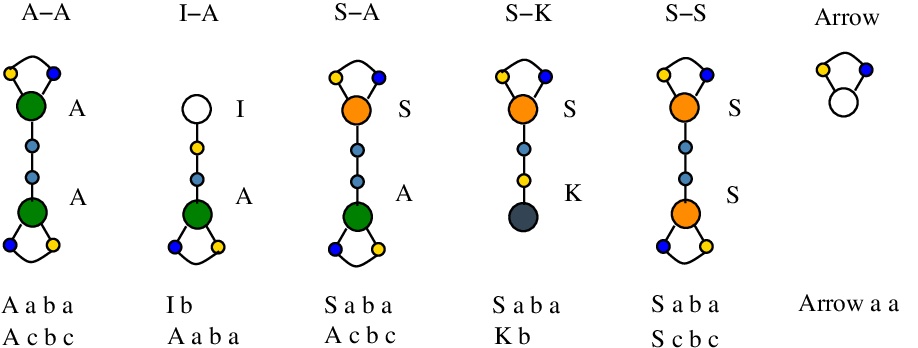}}
\vspace{.5cm}

For each chemSKI rewrite, section \ref{ChemskiRewrites},  are specified the LHS and RHS appear as mol files (with "," as line separator). Moreover, with the addition of tokens there will be two bigger mol files, called "LHS with tokens" and "RHS with tokens". Written as strings (ie with line and field separator equal), "RHS with tokens" will be a permutation of "LHS with tokens".

Borrowing from the \cite{hapax} \href{https://github.com/mbuliga/hapax}{Project Hapax}, we see that the addition of tokens solves the problem of unique name edges, even in a decentralized graphical reduction. It is enough to have "minted" tokens (ie tokens with unique edge names) and then to use them in the rewrites. The minting is of course an independent part from the reduction rewrites.

The tokens have 1 (for Arrow token), 2 (for S-K and I-A tokens) or 3 (for A-A, S-A, S-S tokens) edge names. Minting such a token is a process where the new, unique name edges of the token are produced. In a sense, each such token carries 1, 2 or 3 unique names.

\section{Cost of rewrites}
\label{OnCost}

Each chemSKI rewrite, section \ref{ChemskiRewrites},  has the form

$\left.\right.$ \newline $\left.\right.$ \newline
LHS + In Tokens = RHS + Out Tokens\newline

Suppose that each token has a scalar cost, per token type, in our case there are numbers  cost[A-A], cost[I-A], cost[S-A], cost [S-K], cost[S-S], cost[Arrow].  In the programs the cost per token is in the TokensCost vector \cite{chemskirepo} \href{https://github.com/mbuliga/chemski/blob/master/js/nodes.js#L23}{[nodes.js line 23]}. Then the cost of a collection of tokens is the sum of costs of tokens.

A reasonable choice is to take the cost of the token as the number of  unique names it carries, but this choice is not unique. In an economy application the cost per token would be decided by trading. In a world building application the cost per token would be tuned bo reflect its scarcity.

As the Arrow token enters in many "COMB" rewrites, which just eliminate the Arrow nodes, it is also reasonable to assign a cost equal to 0 for this token, because Arrow nodes in graphs have a fleeting existence.

The cost of a rewrite depends on the application. There is an in cost of a rewrite:

$\left.\right.$ \newline $\left.\right.$ \newline
InCost[rewrite] = cost[In Tokens]  \newline

$\left.\right.$ \newline
and an out cost of a rewrite:

$\left.\right.$ \newline $\left.\right.$ \newline
OutCost[rewrite] = cost[Out Tokens] \newline

$\left.\right.$ \newline
From here, we may have a cost of a rewrite, as 

$\left.\right.$ \newline $\left.\right.$ \newline
cost[rewrite] =  OutCost[rewrite] - InCost[rewrite]\newline

$\left.\right.$ \newline
The cost of a graph reduction would be then the sum of the costs of rewrites performed.

For example, in the case when the cost measures the number of new names, this is reasonable. Also, this cost[rewrite] would be interesting in relation to world building applications, where we see graphs as artificial life, \cite{buligaalife} \href{https://arxiv.org/abs/2005.06060}{arXiv:2005.06060} section 3, or the experiment page \cite{quinecheck} \href{https://mbuliga.github.io/quinegraphs/quinecheck.html}{How to test a quine}. For such an application cost[rewrite] is related to the metabolism.

For economy applications we may use InCost[rewrite] to bill a rewrite. We would use perhaps only "In Tokens" from the customer.  Then OutCost[rewrite] would measure the value of the tokens we obtain because we performed the rewrite. cost[rewrite] would be then the profit.

Also for economy or general computation applications, such cost would supplement other costs considered, like the number of computational steps performed by a virtual machine to do the rewrite. Indeed, for some quine graphs, like in \href{https://mbuliga.github.io/chemski/chemski-with-tokens.html#PlayPlace}{[Play with chemSKI section]} menu the graph associated to (S I I) (S I I), the cost of the graph reduction oscilates in time around 0. The computational cost of running the reductions for such a quine should be something which increases approximately linear in time. For other graphs,  like in \href{https://mbuliga.github.io/chemski/chemski-with-tokens.html#PlayPlace}{[Play with chemSKI section]} menu the graph associated to (S I I) (S (K (S I I)) (S (K (S I)) (S (K K) I))), they may evolve into a "dirty" quine graph, which is an ever increasing graph which has a component which is a quine, but also more and more "dirt"  or "waste" composed of small graphs which are inert. In such a case the cost proposed here is linear in time.

The existence of "waste" implies that, whatever cost we consider, it should also extend to the cost[Initial graph] and cost[Final Graph], such that it satisfies:

$\left.\right.$ \newline $\left.\right.$ \newline
cost[Initial Graph] + InCost[all rewrites performed] = cost[Final Graph] + OutCost[all rewrites performed]\newline

$\left.\right.$ \newline $\left.\right.$ \newline
OutCost[all rewrites performed] = Profit[all rewrites performed] + cost[waste]\newline

We leave open this subject.

\section{chemSKI rewrites}
\label{ChemskiRewrites}

The list of rewrites of chemSKI is this section, in mol notation and graphically.

The rewrites are defined  in \cite{chemskirepo} \href{https://github.com/mbuliga/chemski/blob/master/js/chemistry.js#L97}{[chemistry.js from line 99]}

$\left.\right.$ \newline $\left.\right.$ \newline
"function chemistry(id) { \newline
switch (id) { \newline
case "CHEMSKI": \newline
... }}"

$\left.\right.$ \newline
by a string like for example for the rewrite I-S:  \cite{chemskirepo} \href{https://github.com/mbuliga/chemski/blob/master/js/chemistry.js#L103}{[chemistry.js line 102]}

$\left.\right.$ \newline $\left.\right.$ \newline
{left:"I",right:"S",action:"terminIS", named:"I-S", kind:"TERMINATION"}\newline

$\left.\right.$ \newline
called in the programs comments "a transform".  The meaning of this string is the following. A rewrite is mathematically described by a pair (LHS,RHS), where LHS and RHS are graphs, in mol notation. In order to use a rewrite we need first to identify the LHS pattern in the graph. For this we visit each node of the graph, ie we visit each line in the mol description of the graph. Once we are at a node n1, we verify if the node type is the one from the "right" field. Then, according to the name of the "action" field, there is an algorithm which tries to visit the graph, going from the node n1 by the edges of the graph. The algorithm tries to identify the node n2, which should have the type of the "left" field. The algorithm describes a path in the LHS, starting from n1, passing by n2, in a way which allows to decide if there is a subgraph which is isomorphic with the LHS pattern.

The "named" field is the name of the rewrite which we use in explanations, following the convention to use the node types  "left"-"right". The field "kind" describes what kind of rewrite we have, for example "TERMINATION" describe rewrites which involve pruning nodes connected to a "termination" node.

Therefore, the important fields used for the rewrites are "left", "right" and "action".

As a comment on the same line as the transform string, there is a pair, like for that rewrite I-S

$\left.\right.$ \newline $\left.\right.$ \newline
"tokenIn: I-A            tokenOut: S-A"\newline

$\left.\right.$ \newline
which indicates to the reader a way to make the rewrite conservative by the use of tokens.

\subsection{The rewrite K-A}
\label{RewriteKA}

 corresponds to the SKI combinators rewrite  K a b -> a.  It is a rewrite of kind "BETA", similar to the beta rewrite (in graphical form) from lambda calculus.

$\left.\right.$ \newline $\left.\right.$ \newline
{left:"K",right:"A",action:"KA", named:"K-A", kind:"BETA"}\newline

It is described in \cite{chemskirepo} \href{https://github.com/mbuliga/chemski/blob/master/js/chemistry.js#L99}{[chemistry.js line 99]}. In mol notation,  is described as: 

$\left.\right.$ \newline $\left.\right.$ \newline
LHS = K 1, A 1 a 2, A 2 b c\newline
RHS = K e, Arrow a c, Arrow b e \newline

$\left.\right.$ \newline
and as a chemical reaction by

\vspace{.5cm}
 
\centerline{\includegraphics[width=0.9\textwidth]{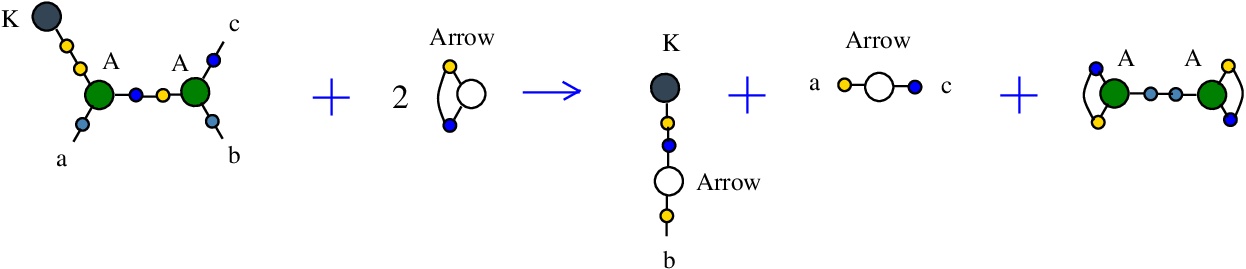}}
\vspace{.5cm}

$\left.\right.$ \newline
with the tokens:  2 Arrow and A-A.

The LHS and RHS patterns with the tokens, are: 

$\left.\right.$ \newline $\left.\right.$ \newline
LHS with tokens = K 1, A 1 a 2, A 2 b c, Arrow d d, Arrow e e\newline   
RHS with tokens = K e, A 1 d 1, A 2 d 2, Arrow a c, Arrow b e\newline

$\left.\right.$ \newline
We can encode the passage from (LHS with tokens) to (RHS with tokens) as the permutation

$\left.\right.$ \newline $\left.\right.$ \newline
(0 1 2 3 4 5 6 7 8 9 a b c d e f)\newline
(0 e 2 1 b 3 6 5 c 7 a 4 9 d 8 f)\newline

\subsection{The rewrite I-A}
\label{RewriteIA}

 corresponds to the SKI combinators rewrite  I a -> a.\newline

$\left.\right.$ \newline
{left:"I",right:"A",action:"termIA", named:"I-A", kind:"TERMINATION"}\newline

$\left.\right.$ \newline
It is described in \cite{chemskirepo} \href{https://github.com/mbuliga/chemski/blob/master/js/chemistry.js#L102}{[chemistry.js line 102]}. In mol notation, is described as: 

$\left.\right.$ \newline $\left.\right.$ \newline
LHS = I 1, A 1 a b\newline
RHS = Arrow a b \newline

$\left.\right.$ \newline
and as a chemical reaction by

\vspace{.5cm}
 
\centerline{\includegraphics[width=0.9\textwidth]{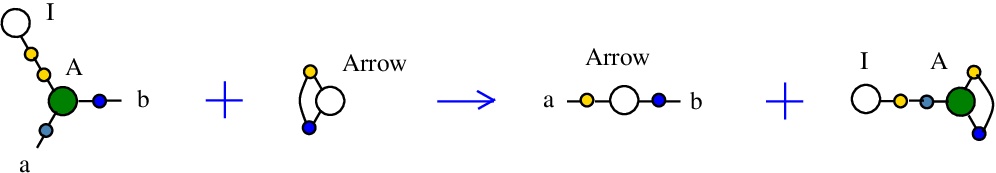}}
\vspace{.5cm}

$\left.\right.$ \newline
with the tokens Arrow and I-A.

The LHS and RHS patterns with the tokens, are: 

$\left.\right.$ \newline $\left.\right.$ \newline
LHS with tokens = I 1, A 1 a b, Arrow c c\newline 
RHS with tokens = I 1, A c 1 c, Arrow a b\newline

We can encode the passage from (LHS with tokens) to (RHS with tokens) as the permutation

$\left.\right.$ \newline $\left.\right.$ \newline
(0 1 2 3 4 5 6 7 8)\newline
(0 1 2 7 3 8 6 4 5)

\subsection{The rewrite I-S}
\label{RewriteIS}

 used to duplicate a I combinator node by the S node which plays the role of a fanout in this rewrite. It can also be seen as a rewrite which eliminates a S node.\newline

$\left.\right.$ \newline
{left:"I",right:"S",action:"terminIS", named:"I-S", kind:"TERMINATION"}

 It is described in \cite{chemskirepo} \href{https://github.com/mbuliga/chemski/blob/master/js/chemistry.js#L103}{[chemistry.js line 103]}. In mol notation, is described as: 

$\left.\right.$ \newline $\left.\right.$ \newline
LHS = I a, S a b c\newline
RHS = I b, I c \newline

$\left.\right.$ \newline
and as a chemical reaction by

\vspace{.5cm}
 
\centerline{\includegraphics[width=0.9\textwidth]{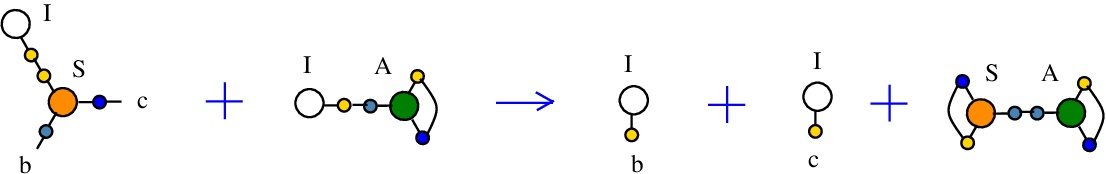}}
\vspace{.5cm}

$\left.\right.$ \newline $\left.\right.$ \newline
with the tokens I-A and S-A.

The LHS and RHS patterns with the tokens, are: 

$\left.\right.$ \newline $\left.\right.$ \newline
LHS with tokens = I a, S a b c, I d, A e d e\newline
RHS with tokens = I b, S a d a, I c, A e d e\newline

$\left.\right.$ \newline
We can encode the passage from (LHS with tokens) to (RHS with tokens) as the permutation

$\left.\right.$ \newline $\left.\right.$ \newline 
(0 1 2 3 4 5 6 7 8 9 a b)\newline
(0 4 2 1 7 3 6 5 8 9 a b)

\subsection{The rewrite K-S}
\label{RewriteKS}

 used to duplicate a K combinator node by the S node which plays the role of a fanout in this rewrite.  

$\left.\right.$ \newline $\left.\right.$ \newline
{left:"K",right:"S",action:"terminKS", named:"K-S", kind:"TERMINATION"}\newline

$\left.\right.$ \newline
It is described in \cite{chemskirepo} \href{https://github.com/mbuliga/chemski/blob/master/js/chemistry.js#L104}{[chemistry.js line 104]}. In mol notation, the  K-S rewrite is described as: 

$\left.\right.$ \newline $\left.\right.$ \newline
LHS = K a, S a b c\newline
RHS = K b, K c

$\left.\right.$ \newline $\left.\right.$ \newline
and as a chemical reaction by

\vspace{.5cm}
 
\centerline{\includegraphics[width=0.9\textwidth]{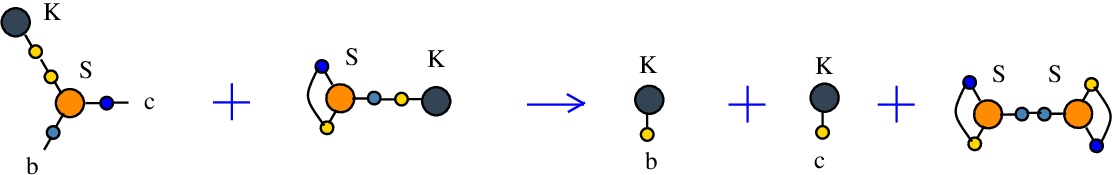}}
\vspace{.5cm}

$\left.\right.$ \newline $\left.\right.$ \newline
with the tokens S-K and S-S.

The LHS and RHS patterns with the tokens, are: 

$\left.\right.$ \newline $\left.\right.$ \newline
LHS with tokens = K a, S a b c, K d, S e d e\newline
RHS with tokens = K b, S a d a, K c, S e d e\newline

$\left.\right.$ \newline 
We can encode the passage from (LHS with tokens) to (RHS with tokens) as the same permutation as the one from the rewrite  I-S 

$\left.\right.$ \newline $\left.\right.$ \newline
(0 1 2 3 4 5 6 7 8 9 a b)\newline
(0 4 2 1 7 3 6 5 8 9 a b)\newline

\subsection{The rewrite S-K}
\label{RewriteSK}

 has two LHS patterns, in both cases the K node, connected at port 2 or port 3 of a S node (when it has the role of a fanout), prunes the respective port.

$\left.\right.$ \newline $\left.\right.$ \newline
{left:"S",right:"K",action:"terminSK", named:"S-K", kind:"TERMINATION"}\newline

$\left.\right.$ \newline 
It is described in \cite{chemskirepo} \href{https://github.com/mbuliga/chemski/blob/master/js/chemistry.js#L105}{[chemistry.js line 105]}. In mol notation, the first S-K rewrite is described as: 

$\left.\right.$ \newline $\left.\right.$ \newline
LHS = S a b c, K c\newline 
RHS = Arrow a b

$\left.\right.$ \newline $\left.\right.$ \newline
and as a chemical reaction by

\vspace{.5cm}
 
\centerline{\includegraphics[width=0.9\textwidth]{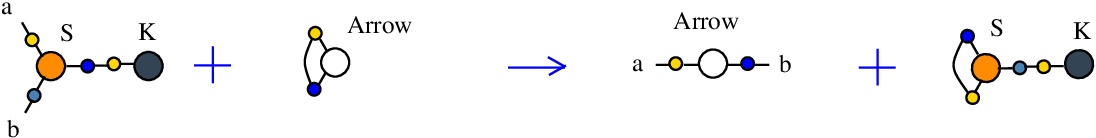}}
\vspace{.5cm}

$\left.\right.$ \newline $\left.\right.$ \newline
with the tokens Arrow and S-L.

The LHS and RHS patterns with the tokens, are: 

$\left.\right.$ \newline $\left.\right.$ \newline
LHS with tokens = S a b c, K c, Arrow d d\newline
RHS with tokens = S d c d, K c, Arrow a b\newline

$\left.\right.$ \newline
We can encode the passage from (LHS with tokens) to (RHS with tokens) as the permutation 

$\left.\right.$ \newline $\left.\right.$ \newline
(0 1 2 3 4 5 6 7 8)\newline
(0 7 3 8 4 5 6 1 2)\newline

The second S-K rewrite is described as: 

$\left.\right.$ \newline $\left.\right.$ \newline
LHS = S a c b, K c\newline 
RHS = Arrow a b

$\left.\right.$ \newline $\left.\right.$ \newline
and as a chemical reaction by

\vspace{.5cm}
 
\centerline{\includegraphics[width=0.9\textwidth]{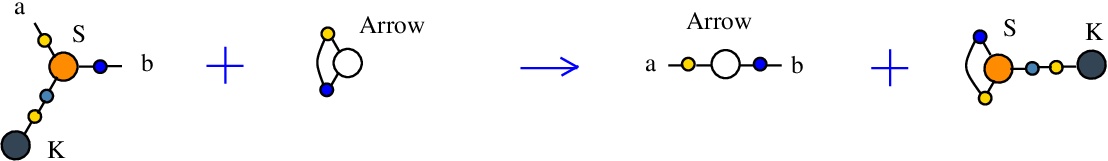}}
\vspace{.5cm}

$\left.\right.$ \newline $\left.\right.$ \newline
with the tokens Arrow and S-K.

The LHS and RHS patterns with the tokens, are: 

$\left.\right.$ \newline $\left.\right.$ \newline
LHS with tokens = S a c b, K c, Arrow d d\newline 
RHS with tokens = S d c d, K c, Arrow a b\newline

$\left.\right.$ \newline
We can encode the passage from (LHS with tokens) to (RHS with tokens) as the  permutation 

$\left.\right.$ \newline $\left.\right.$ \newline
(0 1 2 3 4 5 6 7 8)\newline
(0 7 2 8 4 5 6 1 3)

\subsection{The rewrite A-K}
\label{RewriteAK}

 is a pruning rewrite where a K combinator node connected to the port 3 of an application A node transforms into a pair of K nodes.

$\left.\right.$ \newline $\left.\right.$ \newline
{left:"A",right:"K",action:"termAK", named:"A-K", kind:"TERMINATION"}\newline

$\left.\right.$ \newline 
It is described in \cite{chemskirepo} \href{https://github.com/mbuliga/chemski/blob/master/js/chemistry.js#L106}{[chemistry.js line 106]}. In mol notation, the rewrite is described as: 

$\left.\right.$ \newline $\left.\right.$ \newline
LHS = A a b c, K c\newline 
RHS = K a, K b

$\left.\right.$ \newline $\left.\right.$ \newline
and as a chemical reaction by

\vspace{.5cm}
 
\centerline{\includegraphics[width=0.9\textwidth]{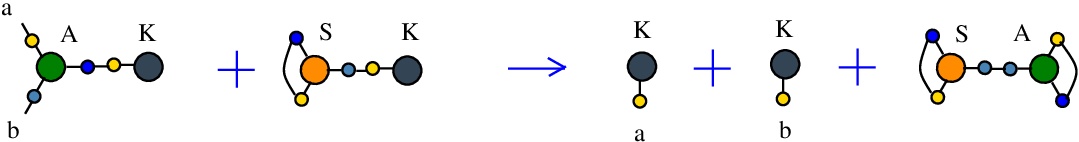}}
\vspace{.5cm}

$\left.\right.$ \newline $\left.\right.$ \newline
with the tokens S-K and S-A.

The LHS and RHS patterns with the tokens, are: 

$\left.\right.$ \newline $\left.\right.$ \newline
LHS with tokens = A a b c, K c, S d e d, K e\newline         
RHS with tokens = A c e c, K a, S d e d, K b\newline

$\left.\right.$ \newline
We can encode the passage from (LHS with tokens) to (RHS with tokens) as the permutation

$\left.\right.$ \newline $\left.\right.$ \newline
(0 1 2 3 4 5 6 7 8 9 a b)\newline
(0 3 8 5 4 1 6 7 8 9 a 2)

\subsection{The rewrite A-S}
\label{RewriteAS}

 duplicates a pair of nodes A and S, thus serving to duplicate A nodes and also S nodes in their role of fanouts.

$\left.\right.$ \newline $\left.\right.$ \newline
{left:"A",right:"S",action:"DIST1", named:"A-S", t1:"S",t2:"S",t3:"A",t4:"A", kind:"DIST"}\newline

$\left.\right.$ \newline
The kind "DIST" takes the name from "distributivity", but I used the name as a generic one for duplication rewrites or even for rewrites which do not increase the number of nodes. The rewrite. However the action field is "DIST1", which is a precise description  because there are 8 types of truly inspired from "distributivity" rewrites (DIST0 to DIST7), as explained in \cite{chemskirepo} \href{https://github.com/mbuliga/chemski/blob/master/js/chemistry.js#L820}{[chemistry.js from line 820]}. In the case of DIST0 to DIST7 rewrites there are new fields "t1", ..., "t4" which are used for the types of nodes in the RHS of the rewrite.

It is described in \cite{chemskirepo} \href{https://github.com/mbuliga/chemski/blob/master/js/chemistry.js#L110}{[chemistry.js line 110]}. In mol notation, the rewrite is described as: 

$\left.\right.$ \newline $\left.\right.$ \newline
LHS = A a b e, S e c d\newline 
RHS = S a e f, S b g h, A e g c, A f h d

$\left.\right.$ \newline $\left.\right.$ \newline
and as a chemical reaction by

\vspace{.5cm}
 
\centerline{\includegraphics[width=0.9\textwidth]{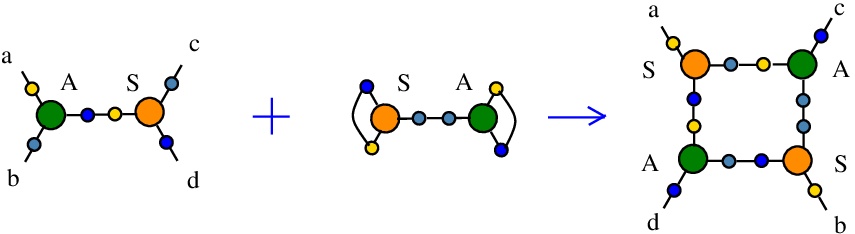}}
\vspace{.5cm}

$\left.\right.$ \newline $\left.\right.$ \newline
with the token  S-A.

The LHS and RHS patterns with the tokens, are: 

$\left.\right.$ \newline $\left.\right.$ \newline 
LHS with tokens = S e c d, S f h f, A a b e, A g h g\newline     
RHS with tokens = S a e f, S b g h, A e g c, A f h d\newline

$\left.\right.$ \newline
We can encode the passage from (LHS with tokens) to (RHS with tokens) as the permutation 

$\left.\right.$ \newline $\left.\right.$ \newline 
(0 1 2 3 4 5 6 7 8 9 a b c d e f)\newline
(0 9 1 5 4 a d 6 8 b f 2 c 7 e 3)

\subsection{The rewrite S-S}
\label{RewriteSS}

 is a neutral rewrite (ie no tokens needed). It uses a S node in the role of an S combinator, connected to another S node in the role of fanout. It duplicates the S combinator, but this is done by rewiring of the two S nodes available.

$\left.\right.$ \newline $\left.\right.$ \newline
{left:"S",right:"S",action:"SS", named:"S-S", kind:"DIST"}\newline

$\left.\right.$ \newline
Mind that even if the kind is "DIST", the action is not one of DIST0 to DIST7, therefore here is a case where I used the kind "DIST" as a placeholder for duplication. This is also justified by the fact that the rewrite does not decrease the number of nodes, it is therefore put in the category of rewrites which is favored by the "GROW" strategy.

This is described in \cite{chemskirepo} \href{https://github.com/mbuliga/chemski/blob/master/js/chemistry.js#L113}{[chemistry.js line 113]}. In mol notation, the rewrite is described as: 

$\left.\right.$ \newline $\left.\right.$ \newline
LHS = S a a b, S b c d\newline 
RHS = S a a c, S b b d

$\left.\right.$ \newline $\left.\right.$ \newline
and as a chemical reaction by

\vspace{.5cm}
 
\centerline{\includegraphics[width=0.9\textwidth]{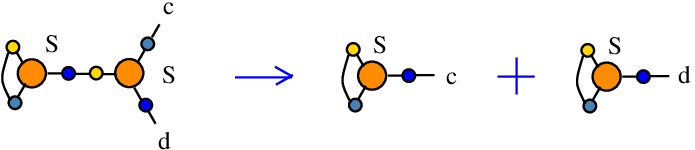}}
\vspace{.5cm}

$\left.\right.$ \newline $\left.\right.$ \newline
with no tokens.

The LHS and RHS patterns with the tokens, are the same, because the rewrite is neutral: 

$\left.\right.$ \newline $\left.\right.$ \newline
LHS with tokens = S a a b, S b c d\newline              
RHS with tokens = S a a c, S b b d\newline

$\left.\right.$ \newline
We can encode the passage from (LHS with tokens) to (RHS with tokens) as the permutation

$\left.\right.$ \newline $\left.\right.$ \newline
(0 1 2 3 4 5 6 7)\newline
(0 1 2 6 4 3 5 7)

\subsection{The rewrite S-A}
\label{RewriteSA}

  is the graphical rewrite version of S a b c -> (a c) (b c). Instead of duplicating "c", it just rewires the LHS where the combinator S appears as an S node with ports 1 and 2 connected, into the RHS where the node S has not the ports 1 and 2 connected, thus it plays the role of a fanout. It is therefore a neutral rewrite (no tokens needed). 

$\left.\right.$ \newline $\left.\right.$ \newline
{left:"S",right:"A",action:"SA", named:"S-A", kind:"DIST"}\newline

$\left.\right.$ \newline
The kind of this rewrite is "DIST", because it does not decrease the number of nodes, therefore it is favored by the rewrite strategy "GROW". It is not a DIST0 to DIST7 rewrite, because the action is "SA".

It is described in \cite{chemskirepo} \href{https://github.com/mbuliga/chemski/blob/master/js/chemistry.js#L114}{[chemistry.js line 114]}. In mol notation, the rewrite is described as: 

$\left.\right.$ \newline $\left.\right.$ \newline
LHS = S 1 1 2, A 2 a 3, A 3 b 4, A 4 c d\newline 
RHS = S c 1 2, A a 1 3, A b 2 4, A 3 4 d

$\left.\right.$ \newline $\left.\right.$ \newline
and as a chemical reaction by

\vspace{.5cm}
 
\centerline{\includegraphics[width=0.9\textwidth]{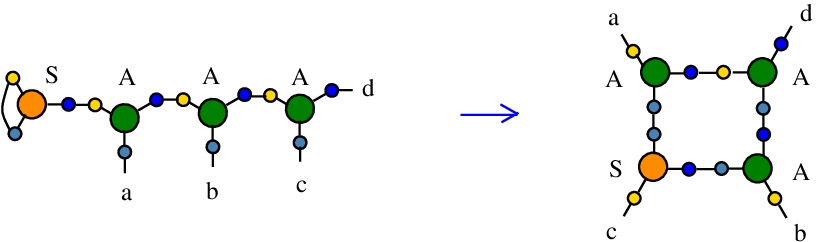}}
\vspace{.5cm}

$\left.\right.$ \newline $\left.\right.$ \newline
with no tokens needed.

The LHS and RHS patterns with the tokens, are the same: 

$\left.\right.$ \newline $\left.\right.$ \newline
LHS with tokens = S 1 1 2, A 2 a 3, A 3 b 4, A 4 c d\newline      
RHS with tokens = S c 1 2, A a 1 3, A b 2 4, A 3 4 d\newline

$\left.\right.$ \newline
We can encode the passage from (LHS with tokens) to (RHS with tokens) as the permutation

$\left.\right.$ \newline $\left.\right.$ \newline
(0 1 2 3 4 5 6 7 8 9 a b c d e f)\newline
(0 e 1 3 4 6 2 7 8 a 5 b c 9 d f)

\subsection{The rewrite COMB}
\label{RewriteCOMB}

 is a generic rewrite which appears in all chemlambda formalisms. It has the role to eliminate Arrow nodes. It appears as  

$\left.\right.$ \newline $\left.\right.$ \newline
{left:"any",right:"Arrow",action:"arrow", named:"COMB", kind:"COMPOSE"}\newline

$\left.\right.$ \newline 
 It is described in \cite{chemskirepo} \href{https://github.com/mbuliga/chemski/blob/master/js/chemistry.js#L15}{[chemistry.js line 15]}. In not quite a mol notation, is described as: 

$\left.\right.$ \newline $\left.\right.$ \newline
LHS = [The other node connected to the edge named "a"], Arrow a b\newline
RHS = [That node where "a" is replaced by "b"] \newline

$\left.\right.$ \newline
It is thus a schema of rewrites. As a a chemical reaction by

\vspace{.5cm}
 
\centerline{\includegraphics[width=0.9\textwidth]{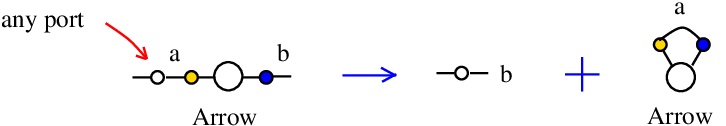}}
\vspace{.5cm}

$\left.\right.$ \newline $\left.\right.$ \newline
with no token as input and a token Arrow as output.

\section{Tokens, synthesis and waste rewrites}
\label{TokensSynthesisRewrites}

Tokens rewrites are rewrites involving only tokens. They are needed in case we have tokens which do not appear as in tokens.

Synthesis rewrites are needed to produce new graphs from old ones, outside of the chemSKI reduction. For example we may want to copy an existing graph, to erase one, or to produce a new graph from existing ones.

Waste rewrites convert waste to tokens or useful graphs.

In this version of the programs these rewrites are not implemented.

\subsection{Tokens rewrites}
\label{TokensRewrites}

  In this proposal we have only one such rewrite:

$\left.\right.$ \newline $\left.\right.$ \newline
S-S + A-A = 2 S-A\newline

The justification is that the tokens A-A and S-S never appear as In Tokens in the  section \ref{ChemskiRewrites}, so there should be a way to convert these tokens into ones which can be used further.

 In mol notation, the rewrite is described as: 

$\left.\right.$ \newline $\left.\right.$ \newline
LHS = S a b a, S c b c, A d e d, A f e f\newline 
RHS = S a b a, S c e c, A d b d, A f e f

$\left.\right.$ \newline $\left.\right.$ \newline
and as a chemical reaction by

\vspace{.5cm}
 
\centerline{\includegraphics[width=0.9\textwidth]{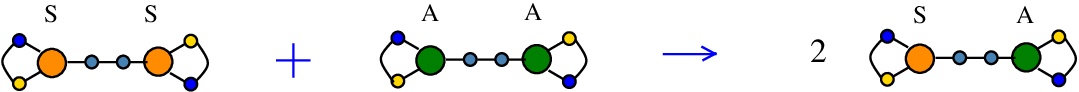}}
\vspace{.5cm}

We can encode the passage from LHS to RHS as the permutation

$\left.\right.$ \newline $\left.\right.$ \newline
(0 1 2 3 4 5 6 7 8 9 a b c d e f)\newline
(0 1 2 3 4 5 a 7 8 9 6 b c d e f)

\subsection{Synthesis S-K rewrite}
\label{SynthesisS-K}

 is a rewrite which rewires two FROUT nodes, with the help of a S-K token. 
In mol notation, the rewrite is described as: 

$\left.\right.$ \newline $\left.\right.$ \newline
LHS with tokens = FROUT a, FROUT b, S c e c, K e\newline 
RHS with tokens = FROUT e, FROUT c, S a e c, K b

$\left.\right.$ \newline $\left.\right.$ \newline
and as a chemical reaction by

\vspace{.5cm}
 
\centerline{\includegraphics[width=0.9\textwidth]{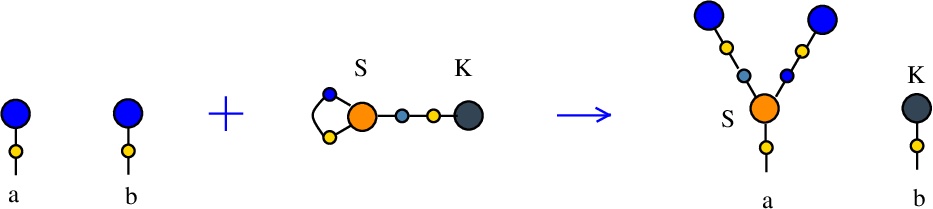}}
\vspace{.5cm}

We can encode the passage from LHS with tokens  to RHS with tokens as the permutation

$\left.\right.$ \newline $\left.\right.$ \newline
(0 1 2 3 4 5 6 7 8 9)\newline
(0 6 2 5 4 1 9 7 8 3)

This rewrite is used to duplicate a graph and erase another. Indeed, suppose FROUT a and FROUT b are the free out nodes corresponding to the roots of two combinators, denoted Ta and Tb (converted to chemSKI). After the rewrite is done, the combinator Ta will be duplicated and the combinator Tb will be erased, by using only the chemSKI rewrites.

These duplication and erasure rewrites will have costs which, in particular, can be used to define the cost of a graph (here cost[Ta]) or to define the cost of erasure of a graph (here the cost of reductions of the Tb graph connected to K). This can be used in section \ref{OnCost}, Cost of rewrites.

In artificial life applications, such a rewrite, applied randomly, will trigger a duplication of a molecule and the erasure of another.

\subsection{Synthesis S-A rewrite}
\label{SynthesisS-A}

 is a rewrite which rewires two FROUT nodes, with the help of a S-K token. In mol notation, the rewrite is described as: 

$\left.\right.$ \newline $\left.\right.$ \newline
LHS with tokens = FROUT a, FROUT b, S c e c, A d e d\newline 
RHS with tokens = FROUT e, FROUT c, S a e d, A d b c

$\left.\right.$ \newline $\left.\right.$ \newline
and as a chemical reaction by

\vspace{.5cm}
 
\centerline{\includegraphics[width=0.9\textwidth]{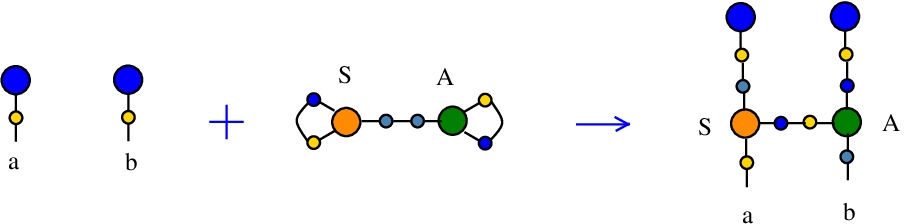}}
\vspace{.5cm}

We can encode the passage from LHS to RHS as the permutation

$\left.\right.$ \newline $\left.\right.$ \newline
(0 1 2 3 4 5 6 7 8 9 a b)\newline
(0 6 2 5 4 1 a 9 8 b 3 7)\newline

$\left.\right.$ \newline
This rewrite is used to duplicate a graph and connect one of the copies to another, via an A node. Indeed, as before suppose FROUT a and FROUT b are the free out nodes corresponding to the roots of two combinators, denoted Ta and Tb (converted to chemSKI). After the rewrite is done, the combinator Ta will be duplicated (by the node S) into Ta1 and Ta2 and  Ta2 will be applied to the combinator Tb, by using only the chemSKI rewrites.

This rewrite may play the role of a rewrite from \cite{krushewski} Combinatory Chemistry: Towards a Simple Model of Emergent Evolution \href{https://arxiv.org/abs/2003.07916}{arXiv:2003.07916}:

$\left.\right.$ \newline $\left.\right.$ \newline
Ta + Tb = Ta Tb

$\left.\right.$ \newline $\left.\right.$ \newline
which does not have a correspondent in chemSKI rewrites.

\subsection{Waste rewrites}
\label{WasteRewrites}

 are rewrites involving tokens and inert 2 nodes graphs, like a pair of K-K  or I-K or I-I  or A-K graphs.

We leave such rewrites unspecified exactly, but it is clear how to propose them. Some could be in a form similar to token rewrites, like

$\left.\right.$ \newline $\left.\right.$ \newline
K-K + S-S = 2 S-K

$\left.\right.$ \newline $\left.\right.$ \newline
I-I + A-A = 2 I-A

$\left.\right.$ \newline $\left.\right.$ \newline
I-K + A-A = A-K + I-A

$\left.\right.$ \newline $\left.\right.$ \newline
A-K + S-S = S-A + S-K

$\left.\right.$ \newline $\left.\right.$ \newline
With each such rewrite comes a cost, building towards a cost[waste], see section \ref{OnCost} Cost of rewrites.

\section{chemSKI +  chemlambda v2}
\label{ChemskiPlusChemlambda}

There is a variant of chemSKI which is fully compatible with chemlambda v2 rewrites, called "chemSKI+ $\lambda$" where the occurences of the node S when there is no edge between ports 1 and 2 are replaced with a chemlambda node FOE. You can toggle between chemSKI and chemSKI+ $\lambda$ by using the "change" button.

You can explore the differences between chemSKI and chemSKI+ $\lambda$ by using the parser window and button " $\lambda$SKI -> mol".

The parser transforms any mix of SKI with lambda calculus (where the letters "S", "K", and "I" are always interpreted as combinators, not lambda calculus variables) into a mol file (i.e. into a graph). This graph can be reduced with chemSKI  or chemSKI plus chemlambda.

In the case of chemSKI the chemlambda nodes do not interpret the node S as a fanout, nor the pure chemSKI nodes (I,S,K) do not see FO and FOE as fanouts. Differently, if you change to the chemistry chemSKI+ $\lambda$ then the reductions work better, with the price of mixing the FO and FOE nodes into the pure chemSKI formalism.

\section{chemSKI + directed Interaction Combinators?}

There is not yet a variant of chemSKI compatible with dirIC \cite{buligaalife} (\href{https://mbuliga.github.io/quinegraphs/ic-vs-chem.html#icvschem}{directed Interaction Combinators}), because in dirIC the rewrite A-FOE from chemlambda is replaced by a rewrite FI-A. This breaks the adaptation chemSKI+ $\lambda$ of chemSKI to chemlambda by replacing the node S as a fanout by a node FOE. It is an interesting problem to consider.

\section{How are chemSKI rewrites used}
\label{ChemskiUsed}

All the programs are abundantly commented, for the convenience of the reader. The way rewrites are done by the programs is described shortly here.

The recognition of the LHS pattern is defined in \cite{chemskirepo} \href{https://github.com/mbuliga/chemski/blob/master/js/chemistry.js#L293}{[chemistry.js from line 293]} in the function 
findTransform(n1), where "n1" is the node where the recognition of the LHS pattern algorithm starts (it should have the type ot the "right" field of the transform). The rewrite is identified by the "action" field, for our example, the rewrite I-S, we have  action:"terminIS", and the recognition of the LHS pattern for that transform is done at 
\cite{chemskirepo} \href{https://github.com/mbuliga/chemski/blob/master/js/chemistry.js#L448}{[chemistry.js lines 448-456]}, which starts with:

$\left.\right.$ \newline $\left.\right.$ \newline
"case "terminIS": case "terminIFOE": "

$\left.\right.$ \newline $\left.\right.$ \newline
(Thus  we may have the same algorithm of recognizing LHS patterns for different action fields, which is not surprising, because the algorithm describes an abstract way to visit a pattern and to perform checks.)

If the LHS pattern of the transform trans (defined by the transform  "action" field and the node n1 seen as a node of the type defined by the "right" field) is found then the function returns the transform trans. Then the graph rewrite is done by the function doTransform(n1, trans), defined in \cite{chemskirepo} \href{https://github.com/mbuliga/chemski/blob/master/js/chemistry.js#L602}{[chemistry.js from line 602]}. At the level of mol files, doing the rewrite means: eliminate the lines in the mol file which correspond to the nodes from the identified LHS pattern, add to the mol file the new lines corresponding to the RHS pattern.

There is no canonical way, no particular order of the lines in the mol file associated to a graph. Any permutation of the lines in the mol file describes the same graph. Any renaming of the edges describes the same graph. That is why we randomly shuffle the mol file before looking for rewrites and we also randomize the choice of rewrites which will be applied. The probability of application of rewrites depends on their "kind", but roughly there is a parameter  which is controlled by the "rewrites weight slider" which favors either the rewrites of the kind which increases (or does not decrease)the number of nodes (ie "GROW"), typically of kind "DIST",  or the kind which decreases the number of nodes (ie "SLIM"), such as kinds "TERMINATION" or "BETA". Another control we have is to associate to the edges of the graph an age (how many time steps since that edge appeared) and then to favor the rewrites performed on older edges first. As the programs don't do rewrites in parallel (although the first version, in awk, \href{https://github.com/chorasimilarity/chemlambda-gui/blob/gh-pages/dynamic/README.md}{chemlambda-gui}, does this for chemlambda v2), this age-based choice of rewrites is a good approximation of the choice to do as much as possible independent rewrites, as soon as possible. This age-based strategy is not used in this version of chemSKI. See \cite{buligaalife} \href{https://mbuliga.github.io/quinegraphs/ic-vs-chem.html}{Alife properties of directed interaction combinators vs. chemlambda.  Marius Buliga (2020)}, \href{https://arxiv.org/abs/2005.06060}{arXiv:2005.06060}.

\section{From SKI to chemSKI}
\label{FromSkiToChemski}

The chemSKI molecule associated to a SKI combinator term is simply the syntax tree of the term, which has as leaves the S,K,I combinator nodes. Go to \href{https://mbuliga.github.io/chemski/chemski-with-tokens.html#PlayPlace}{[Play with chemSKI section]} to use the parser. Use the window " $\lambda$  or SKI -> mol" to parse a SKI combinator to chemSKI. Actually this is a parser from lambda calculus AND SKI combinators to chemlambda AND chemSKI, where the letters "S", "K" and "I" are treated as in SKI. The source is at \cite{chemskirepo} \href{https://github.com/mbuliga/chemski/blob/master/js/0parser.js}{[0parser.js]}. \newline

Examples:\newline
SII will not work, use a space for application: S I I will be understood as the term (S I) I. \newline
(S I I) (S I I) will work\newline
Also, the parser works  for lambda terms, like ($\backslash$x.x x) ($\backslash$x.x x) or  ($\backslash$x.$\backslash$y.x) z.

\section{Play with chemSKI}
\label{PlayPlace}

The github online version of this article allows to experiment with chemSKI. Available here: \href{https://mbuliga.github.io/chemski/chemski-with-tokens.html#PlayPlace}{[Play with chemSKI section]}

\end{document}